\newcommand{\E}{\mathrm{E}}
\documentclass{article}

\usepackage[preprint]{neurips_2021}

\usepackage[utf8]{inputenc} 
\usepackage[T1]{fontenc}    
\usepackage{hyperref}
\hypersetup{urlcolor=cyan}
\usepackage{url}            
\usepackage{booktabs}       
\usepackage{amsfonts}       
\usepackage{nicefrac}       
\usepackage{microtype}      
\usepackage{xcolor}         

\usepackage{graphicx}
\usepackage{amsmath}
\usepackage{amssymb}
\usepackage{microtype}

\title{HAC Explore: Accelerating Exploration with Hierarchical Reinforcement Learning}

\author{%
  Willie ~McClinton\thanks{Equal contribution, Correspondence to wbm3@mit.edu}~~\thanks{Work done as a member of the Google AI Residency Program (g.co/airesidency)} \\
  MIT, CSAIL \\
  \texttt{wbm3@mit.edu} \\
  
  \And

  Andrew ~Levy$^*$ \\
  Brown University\\
  \texttt{levy2@brown.edu} \\
  
  \And

  George ~Konidaris \\
  Brown University\\
  \texttt{gdk@cs.brown.edu} \\
  }

\begin{document}

\maketitle

\begin{abstract}
  Sparse rewards and long time horizons remain challenging for reinforcement learning algorithms. Exploration bonuses can help in sparse reward settings by encouraging agents to explore the state space, while hierarchical approaches can assist with long-horizon tasks by decomposing lengthy tasks into shorter subtasks.  We propose \textit{HAC Explore} (HACx), a new method that combines these approaches by integrating the exploration bonus method Random Network Distillation (RND) into the hierarchical approach Hierarchical Actor-Critic (HAC).  HACx outperforms either component method on its own, as well as an existing approach to combining hierarchy and exploration, in a set of difficult simulated robotics tasks.  HACx is the first RL method to solve a sparse reward, continuous-control task that requires over 1,000 actions.  
\end{abstract}

\section{Introduction}

Tasks that involve long time horizons and sparse rewards remain difficult for reinforcement learning (RL) algorithms.  In the recent literature, two promising approaches have emerged to address these domains: (i) hierarchical reinforcement learning (HRL) \citep{levy2018hierarchical, NEURIPS2018_e6384711,eysenbach2018diversity,Bagaria2020Option, DBLP:journals/corr/abs-1910-11956,ajay2020opal, zhang2021hierarchical} and (ii) exploration bonus methods \citep{ NIPS2016_afda3322,pathak2017curiosity-driven,burda2018exploration}.  

Yet these approaches only address one aspect of these challenging domains---either long time horizons or sparse rewards---but not both.  
HRL methods have improved sample efficiency in domains that require learning a long sequence of primitive actions because they can decompose the problem into relatively short subpolicies.  Yet many of these methods avoid the exploration problem of reaching distant rewarding states by incorporating handcrafted exploration tricks such as large initial state spaces, dense rewards, and demonstrations.  On the other hand, exploration bonuses make it easier for agents to explore new states in sparse reward tasks by integrating intrinsic rewards; but the empirical successes of exploration bonus methods have largely been in discrete action domains rather than long horizon, continuous action domains.    
There have been some attempts to combine hierarchy and exploration \citep{eysenbach2018diversity, roder2020curious, zhang2021hierarchical}, but these methods are limited due to their inability to learn multiple levels of policies efficiently in parallel, and weak exploration incentives.

\begin{figure*}[h]
    \centering
    \includegraphics[width=1.0\linewidth]{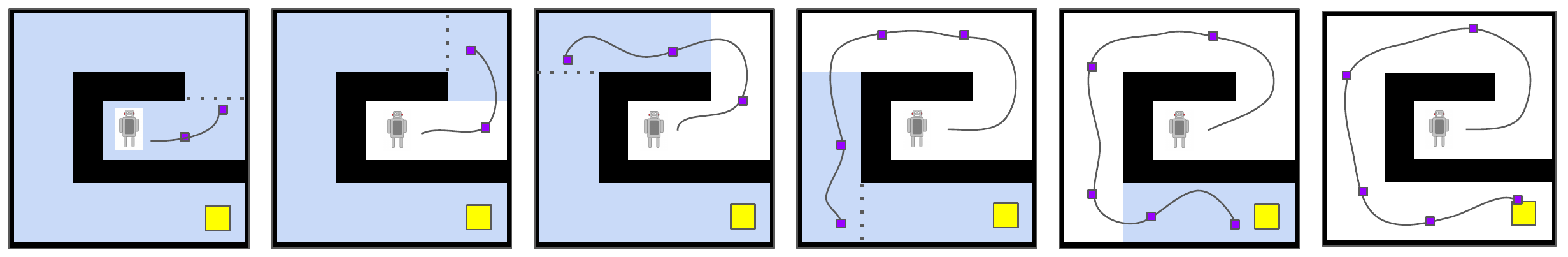}
    \caption{An illustration of the hierarchical exploration policy learned during each stage of the RND curriculum.  At the beginning of each stage, RND identifies the regions of the state space that have not been frequently visited (blue shaded regions), returning a sparse reward if the agent enters these goal regions.  During each stage of training, the agent learns a $k$-level HAC policy (in this case, $k=2$) that tries to decompose the task of reaching the goal regions.  In this example, the top level learns a short sequence of subgoal states (purple squares) to reach the blue regions.  At the same time, the bottom level learns a short sequence of atomic actions to reach each subgoal.  Using the transitions generated by the hierarchical exploration policy, the agent also learns a hierarchical goal-conditioned policy capable of reaching specific target states like the yellow square.}
    \label{fig:hacx_toy_ex}
\end{figure*}

We introduce a new way to combine hierarchy and exploration bonuses that enables agents to explore over long distances in continuous domains.  Our approach, HAC Explore (HACx), integrates a variant of the Random Network Distillation (RND) \citep{burda2018exploration} exploration bonus method into the Hierarchical Actor-Critic (HAC) \citep{levy2018hierarchical} HRL algorithm.  
RND directly encourages exploration by generating a curriculum of subproblems for the agent, each of which incentivizes the agent to further expand the frontier of visited states.  At the beginning of each stage of the curriculum, the RND component identifies the state regions that have not been thoroughly visited, and provides a sparse reward if the agent reaches those regions during training.  

The HAC component of the algorithm learns two hierarchical policies: (i) a hierarchical exploration policy and (ii) a hierarchical goal-conditioned policy.  The purpose of the hierarchical exploration policy is to accelerate exploration over long distances by decomposing the problem of reaching new states into short-horizon subpolicies.  Figure \ref{fig:hacx_toy_ex} shows a toy example of the hierarchical exploration policy learned during each stage of the RND curriculum.  The purpose of the hierarchical goal-conditioned policy is to learn to break down the task of reaching particular states, using the experience transitions generated by the hierarchical exploration policy.  HAC was chosen as the HRL component because, unlike other HRL methods, HAC can break down a task among the levels of the agent so that each level can solve its easier subtask in parallel with the other levels. 

We evaluated HACx in a set of simulated robotics domains, designed in MuJoCo \citep{conf/iros/TodorovET12}.  In the sparse reward, long horizon continuous-control tasks, HACx substantially outperforms HAC, RND, and Curious HAC (CHAC) \citep{roder2020curious}, another method that combines HRL and exploration bonuses.  To our knowledge, HACx is the first algorithm to successfully solve a sparse reward, goal-conditioned task in a continuous domain that requires over 1,000 primitive actions.\footnote{A video presentation of our experiments is available at \url{https://www.youtube.com/watch?v=o9sbAZOtFAQ}.}

\section{Background}


We consider Universal Markov Decision Processes (UMDPs) (or goal-conditioned MDPs)  \citep{pmlr-v37-schaul15, levy2018hierarchical}, in which an agent must learn to reach a given goal region.  A UMDP is defined by the tuple ($\mathcal{S},\mathcal{G},\mathcal{A},T,R,\phi_s,\phi_g,\gamma$), where $\mathcal{S}$ is the state space; $\mathcal{G}$ is the goal space, which is typically the full state space $\mathcal{S}$ or a function of the state space $g(\mathcal{S})$ (e.g., certain dimensions of the state space); $\mathcal{A}$ is a continuous action space;  $T: \mathcal{S}_t \times \mathcal{A}_t \times \mathcal{S}_{t+1} \rightarrow [0,1]$ is the transition dynamics, giving the probability of the next state conditioned on the prior state and action;  $R: g(\mathcal{S}) \times \mathcal{G} \rightarrow \mathbb{R}$ is the reward function that will generate a sparse reward if $g(s_t)$ ($g(\cdot)$ is the function that maps the state space to the goal space) is within an $\epsilon$-ball of the episode goal $g$;  $\phi_s$ and $\phi_g$ are the initial state and goal distribution, respectively; and  $\gamma: \mathcal{S} \rightarrow \{0.99, 0\}$ is the discount rate function.  We assume the episode will terminate (i.e. $\gamma = 0$) if the current state is within the $\epsilon$-ball of the goal.  The objective in a Universal MDP is to learn a goal-conditioned policy $\pi_{\theta}(\cdot|s_t,g)$---a probability distribution over the action space conditioned on the current state and goal---that maximizes the expected, cumulative, and discounted reward.

\subsection{Random Network Distillation}

Random Network Distillation is an exploration bonus method that provides agents with additional rewards for reaching less frequently visited states.  In the MDP setting, RND uses a simple way to measure how often a state has been visited: progress on a supervised learning problem mapping states to random codes.  The exploration reward provided to the agent is proportional to the error on the supervised learned problem.          

More formally, RND involves two randomly initialized neural networks: a fixed target network $\hat{f}: h(\mathcal{S}) \rightarrow \mathbb{R}^d$ and a predictor network with trainable parameters $\theta$ $f_{\theta}: h(\mathcal{S}) \rightarrow \mathbb{R}^d$  that map states or functions of states $h(s)$ to $d$-dimensional codes.  The predictor network is optimized via stochastic gradient descent to minimize the mean squared error between the predicted and targeted codes with the batch of states sampled from buffer $\beta$:
\begin{equation*}
\min\limits_{\theta} \E_{s_t \sim \beta}[(f_{\theta}(s_t) - \hat{f}(s_t))^2].
\end{equation*}
RND sets the exploration reward to be proportional to the squared error in the predicted code as states that have been seen frequently should have lower errors.   The RL objective in RND is to maximize the sum of the task and exploration rewards.  
RND has shown state-of-the-art performance in Montezuma’s Revenge and similar hard exploration Atari games \citep{burda2018exploration}, but  has not shown that it can help agents explore over long distances in domains with continuous actions.

\subsection{Hierarchical Actor-Critic}
\label{HAC section}

Hierarchical Actor-Critic is a method within the feudal class of hierarchical RL algorithms.  As a feudal HRL algorithm, HAC implements agents that contain $k$ levels of policies $\pi_0,\pi_1,\dots,\pi_{k-1}$.  Policies $\pi_1,\pi_2,\dots,\pi_{k-1}$ take as input the current state and the goal state from the level above (the top level policy $\pi_{k-1}$ receives a goal sampled from the task goal space) and output subgoal states for the level below to achieve.  The base level policy $\pi_0$ uses the same structure but outputs primitive actions.

The key benefit of HAC is that the algorithm helps hierarchical agents more efficiently learn multiple levels of policies in parallel because it can overcome the problem of nonstationary transition and reward functions.  In a typical hierarchical agent, the transition and reward functions at any level depend on the policy from the next level below.  This dependency makes learning multiple levels of policies at the same time difficult because changing policies at one level can produce nonstationary transition and reward functions at all higher levels.  Nonstationary transition and reward functions render RL methods less effective as they make it hard to estimate the value of an action.  

HAC can overcome this problem of unstable transition and reward functions by valuing value subgoal actions as if lower level policies are already optimal and fixed instead of valuing a subgoal action based on the current and potentially unstable lower level policy hierarchy.  HAC can train a subgoal level in this manner without actually needing access to the optimal lower level policies primarily as a result of a simple manipulation of the transitions the subgoal levels receive.  When a subgoal level proposes a subgoal, but the level below is unable to achieve the subgoal after $H$ actions, the level that proposed the original subgoal action receives a transition in which the action component is the subgoal state reached in hindsight after $H$ actions, not the original subgoal.  The action and next state components in the transition are now the same, mimicking how an optimal lower level policy would have acted.  

More formally, the subgoal levels in HAC operate in a UMDP that is a weighted average of two UMDPs.  We refer to these two UMDPs as the Future UMDP and the Present UMDP, and they  can be defined by the following tuples.
\begin{enumerate}
    \item Future UMDP - ($\mathcal{S}^i,\mathcal{G}^i,\mathcal{A}^i,T_{\Pi_{i-1}^*},R_{\Pi_{i-1}^*},\phi_{s}^i,\phi_{g}^i,\gamma^i$)
    \item Present UMDP - ($\mathcal{S}^i,\mathcal{G}^i,\mathcal{A}^i,T_{\Pi_{i-1}},R_{\Pi_{i-1}},\phi_{s}^i,\phi_{g}^i,\gamma^i$)
\end{enumerate}

The key differences in these UMDPs are the transition and reward function components.  The purpose of the Future UMDP is to train a subgoal level as if the lower level policy hierarchy has already reached its future, optimal form in order to avoid instability issues.  Accordingly, the transition function at level $i$ is denoted $T_{\Pi_{i-1}^*}$ as it uses the optimal lower level policy hierarchy $\Pi_{i-1}^* = \{\pi_0^*,\dots,\pi_{i-1}^*\}$ instead of the current lower level policy hierarchy $\Pi_{i-1}= \{\pi_0,\dots,\pi_{i-1}\}$.  The Future UMDP does not require access to optimal lower level policy hierarchy $\Pi_{i-1}^*$ because it can instead sample from $\Pi_{i-1}^*$ using the simple subgoal replacement strategy discussed earlier.  In order for the reward to reflect the optimal lower level policies without actually needing access to them, the sparse reward function $R_{\Pi_{i-1}^*}$ only depends on the state reached by the subgoal action and the goal state.  Specifically, HAC uses the reward function $R_{\Pi_{i-1}^*} = 0$ if $ |s_t - g| < \epsilon$ (i.e., the agent is within an $\epsilon$-ball of the goal) and $R_{\Pi_{i-1}^*} = -1$ otherwise.  The Future UMDP thus encourages each subgoal level to learn the shortest path of subgoal states to a goal state, regardless of whether the subgoals can be achieved by the current lower level policies.  

The Future UMDP is not sufficient to train multiple levels of policies in parallel in continuous domains because it does not produce any transitions for the large space of subgoal states that are beyond the reach of the level below in $H$ actions.  HAC uses transitions generated from the Present UMDP to overcome this difficulty.  Transitions are generated from the Present UMDP using a procedure known as subgoal testing.  A certain percentage of the time after a subgoal is proposed (e.g., the original HAC paper used 30\%), the lower policy hierarchy is followed exactly (i.e., the transition function is $T_{\Pi_{i-1}}$).  If the proposed subgoal is achieved, then the same reward function is used as the Future UMDP.  However, if the subgoal is not achieved, a low reward is given to the agent (e.g., HAC uses a reward of $-H$, the opposite of the maximum number of actions level $i$ has to achieve its goal).  The Present UMDP thus encourages level $i$ to learn the shortest path of currently achievable subgoal states that lead to the goal state.  Yet since a subgoal level in HAC receives significantly fewer transitions from the Present UMDP relative to the Future UMDP, the weighted average of the two UMDPs primarily encourages the agent to learn the shortest path of subgoals that both can be achieved by the optimal lower level policy hierarchy $\Pi_{i-1}^*$ and lead to the goal state, but now the overly distant subgoals are discouraged.

Although HAC has shown that it can achieve better sample efficiency than existing HRL methods in long horizon tasks, a weakness in the algorithm is that it does not directly incentivize agents to explore new states.  As a result, when HAC is used in tasks in which the goal states are far from the initial state space, the algorithm breaks down because the top level subgoal policy receives no signal as to where subgoals should be placed.  The original HAC experiments sidestepped this problem by setting the initial state space to the full state space, which served as a hand-crafted curriculum for the agent to reach new states.  However, this strategy makes a strong assumption that the agent can be placed anywhere in the state space.  Further, the strategy is unlikely to scale to very large domains in which the sampled initial state is still far from the goal state.

\section{HAC Explore}

To improve the sample efficiency of HAC in environments that require exploration over long distances, we propose HAC Explore.  HAC Explore helps agents explore over long time horizons by integrating the RND exploration bonus method into HAC. RND generates a curriculum of subproblems that each encourage the agent to move into regions of states that have not been frequently explored.  The HAC component then accelerates exploration over long distances by dividing up the task of reaching distant new states into a parallel set of easier subtasks.

\begin{figure}
    \centering
    \includegraphics[width=0.55\linewidth]{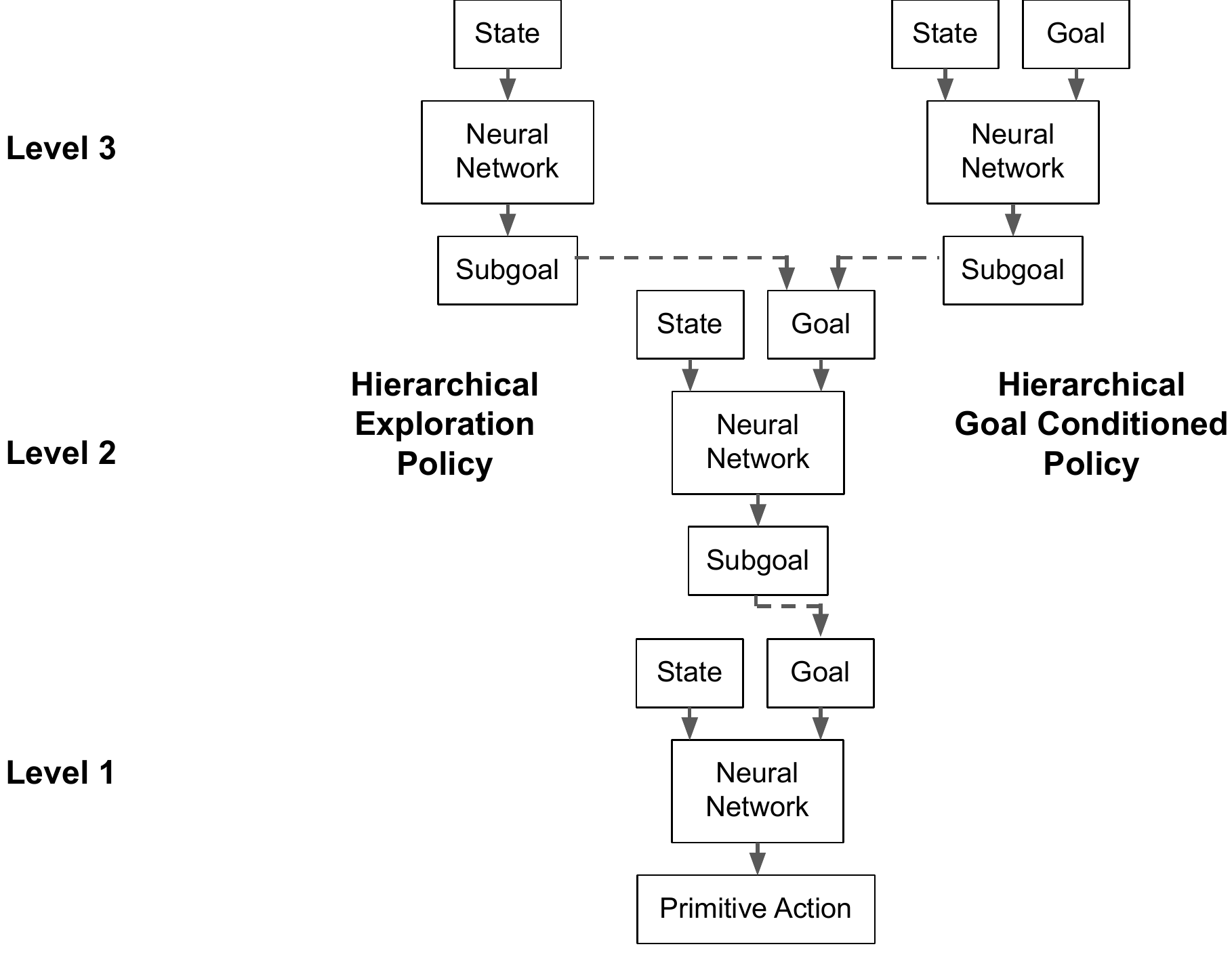}
    \caption{The architecture of the two hierarchical policies learned by a 3-level HACx agent.  The top level has two different subgoal policies: (i) an exploration policy that seeks to learn a sequence of subgoals to reach new regions of the state space and (ii) a goal-conditioned policy that seeks to learn a sequence of subgoals to reach particular states.  The bottom two levels are shared between the hierarchical policies.}
    \label{fig:hacx_arch}
\end{figure}

\subsection{Separation of Exploration and Goal Reaching}

Agents in sparse reward and long time horizon UMDPs must both explore new areas and learn to reach particular states.  Combining these two objectives into a single reward function can be impractical for two reasons.  First, significant domain expertise would likely be required to shape the combined reward function such that both objectives are met rather than one dominating the other.  For instance, a designed would have to carefully tune a parameter than weights the task reward against the exploration bonus, to encourage exploration but not damage performance.  Second, when the goal-conditioned policy must be tested, it  may be unable to reach the goal even though the agent may have sufficient experience to do so, because the reward function diverts the policy to explore new areas.

Instead of combining these objectives into one reward function, HACx treats exploration and goal reaching as separate problems.  Accordingly, HACx agents learn two separate hierarchical policies.  The hierarchical exploration policy seeks to break down the task of reaching new regions of the state space, while the hierarchical goal-conditioned policy is the regular HAC policy that seeks to decompose the task of reaching particular states.  Figure \ref{fig:hacx_arch} shows the architecture of a 3 level HACx agent.  The top level of HACx agents has two policies: (i) an exploration subgoal policy $\pi_e: \mathcal{S} \rightarrow \mathcal{S}$ and a goal-conditioned policy $\pi_g: \mathcal{S} \times \mathcal{G} \rightarrow \mathcal{S}$.  The exploration policy $\pi_e$ takes as input the current state and outputs a subgoal state.  The goal-conditioned policy $\pi_g$ takes as input the current state and goal and outputs a subgoal state.  The two hierarchical policies share the remaining lower level goal-conditioned policies that  further break down subgoals given by the top levels.

\subsection{Subgoal Exploration MDPs}

Similar to HAC, HACx takes as input a task UMDP ($\mathcal{S},\mathcal{A},T,R,\phi_s,\phi_g,\gamma$) and transforms it into a hierarchy of UMDPs.  The key difference is the addition of exploration MDPs at the top level.  Like the subgoal policies in HAC, the subgoal exploration policy in HACx seeks to maximize reward in a weighted average of a Future Exploration MDP and Present Exploration MDP, which can be defined with the following tuples:
\begin{enumerate}
    \item Future Exploration MDP - ($\mathcal{S},\mathcal{A},T_{\Pi_{i-1}^*},R_{\Pi_{i-1}^*},\phi_s,\gamma$)
    \item Present Exploration MDP - ($\mathcal{S},\mathcal{A},T_{\Pi_{i-1}},R_{\Pi_{i-1}},\phi_s,\gamma$)
\end{enumerate}

The key differences from the Future and Present UMDPs in HAC are the reward functions.  The subgoal exploration policy is not trying to break down the task of reaching a particular state as in HAC, but rather trying to divide the problem of reaching new regions in the state space.  As a result, the reward function for the Future and Present Exploration MDPs provides a sparse reward if the agent has entered a ``new'' state.  

HACx uses a strategy similar to RND to determine whether a state is ``new''.  As in RND, HACx agents have both learned RND code $f_{\theta}: h(\mathcal{S}) \rightarrow \textrm{R}^d$ and true RND code $\hat{f}: h(\mathcal{S}) \rightarrow \textrm{R}^d$ functions that map states or functions of states $h(\mathcal{S})$ to $d$-dimensional codes.  HACx will classify a visited state $s_t$ as ``new'' if the learned RND code $f_{\theta}(s_t)$ is not within an $\epsilon$-ball of the true RND code $\hat{f}(s_t)$ because the less an agent has visited a state the worse the learned RND code mapping should be at predicting the true RND code.  That is, in both the Future and Present Exploration MDPs, $R = 0$ if $|f_{\theta}(s_t) - \hat{f}(s_t)| > \epsilon$ else $R = -1$.  Additionally, the subgoal exploration policy will still be penalized with a reward of $-H$ if proposed subgoals are missed during subgoal testing.  Consequently, the Future and Present exploration MDPs, encourage the subgoal policy to learn the shortest sequence of subgoal states that both can be achieved by the optimal lower level policy hierarchy $\Pi_{i-1}^*$ and lead to regions of the state space that have not been frequently visited. 

HACx updates the RND exploration reward in a curriculum-like manner.  For a fixed number of episodes, the learned RND network is not updated in order to ease value estimation by making rewards more stationary. After the batch of episodes is completed, a large update is made to the learned RND code mapping, resulting in a smaller region of ``new'' states for the exploration policy to learn to reach.  

\subsection{Behavior Policy and Base RL Algorithm}

HACx agents and their two hierarchical policies interact with the environment using the following behavioral policy.  After an initial state is sampled from the distribution $\phi(s)$, a certain fraction of the time $\tau$, the top level exploration subgoal policy is randomly chosen to propose a subgoal (in our experiments, $\tau = 0.6$).  The remaining fraction of time, the top level goal-conditioned policy is selected.  After the top level policy proposes a subgoal, control passes to the goal-conditioned policy at the next level down, which then has $H$ actions to achieve the goal provided by the level above.  To aid exploration, noise is also added to all actions when subgoal testing is not occurring.  Specifically, HACx applies diagonal Gaussian noise $\mathcal{N}(0,\sigma^2)$ to the actions prescribed by the policies.  Higher level policies can act at larger time scales so a higher variance $\sigma^2$ of noise is added to policies that are higher in the policy hierarchy.  The extra noise at higher levels  may further accelerate exploration for HACx agents as it  widens  the range of states that can be realistically reached.

All policies within a HACx agent are trained using Deep Deterministic Policy Gradient (DDPG) \citep{journals/corr/LillicrapHPHETS15} without target networks as the base RL algorithm.  All policies are thus deterministic and trained with the help of a parameterized Q-function $q_{\phi}: \mathcal{S} \times \mathcal{G} \times \mathcal{A} \rightarrow \mathbb{R}$ and a replay buffer $\beta$ that will store $(s_t,a_t,r_{t+1},s_{t+1},g_t \text{ (if UMDP)},\gamma)$ tuples.  Updates to the policy and Q-function are interleaved.

\subsection{HACx Example and Full Algorithm}

Please refer to the supplementary material for (i) a toy example that discusses in detail how the HACx behavior policy works and the transitions that are generated in an episode and (ii) the full HACx algorithm.

\begin{figure*}
    \centering
    \includegraphics[width=0.9\linewidth]{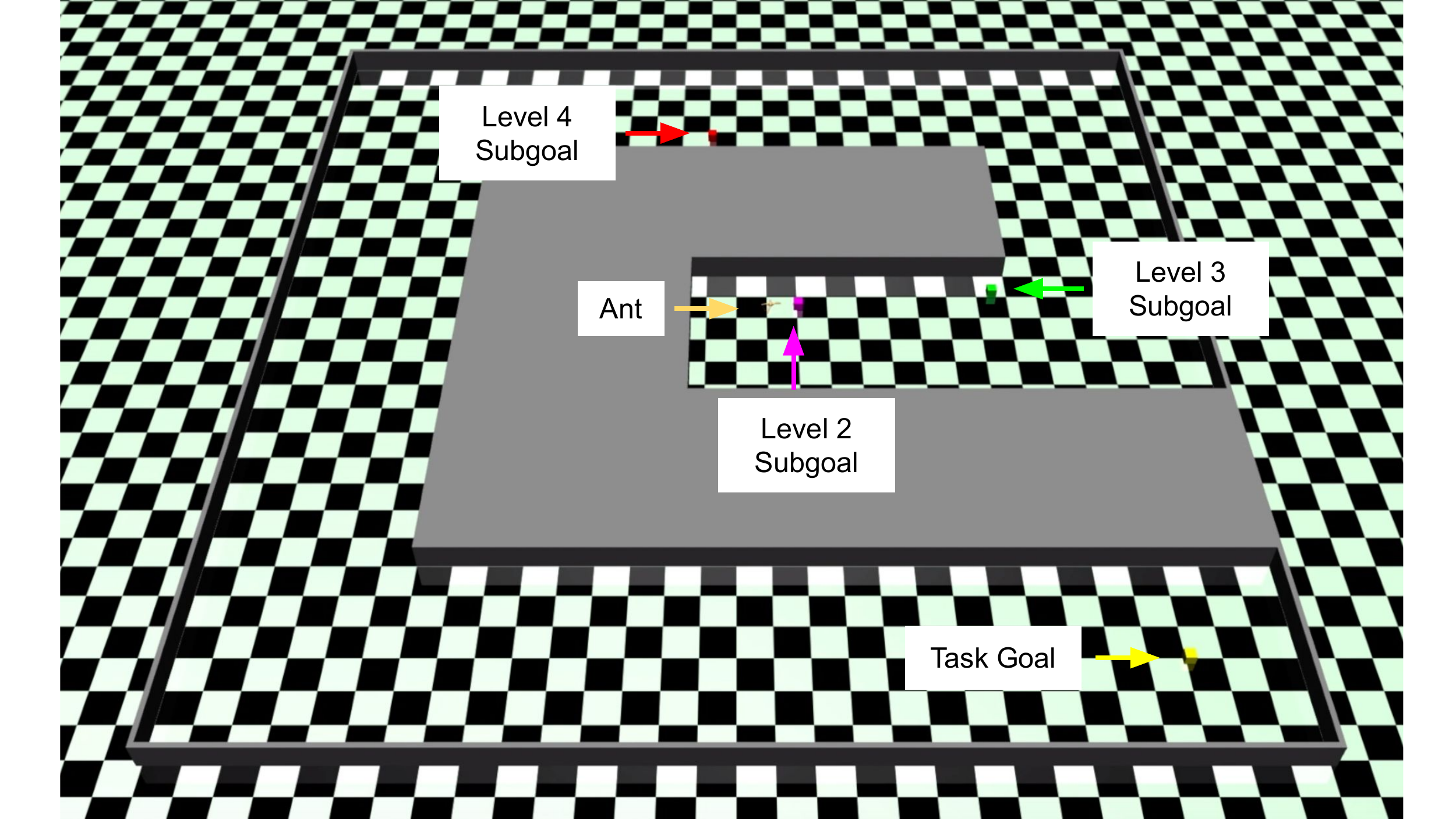}
    \caption{Figure shows a trained 4-level HACx agent in the spiral maze domain.  The goal of the task is to move from a small region in the center of the maze to the bottom right corner of the maze (in this case, the yellow cube).  The figure also shows the initial subgoals proposed by each level.  In order to reach its goal of the yellow cube, level 4 of the agent proposes the subgoal state shown by the red cube.  In order to achieve its goal of the red cube, level 3 proposed the subgoal state shown by the green cube.  In order to achieve its goal of the green cube, level 2 proposes the subgoal shown by the purple cube.}
    \label{fig:spiral_maze}
\end{figure*}

\section{Experiments}

We evaluated HACx on a set of three simulated robotics domains built with MuJoCo \citep{conf/iros/TodorovET12}.  Two of the tasks require over an order of magnitude more primitive actions than the longest domains in the HAC paper.  Further, none of the environments use large initial state and goal spaces to help with exploration, as was done in the original HAC paper.  A video presentation of our experiments is available at \url{https://www.youtube.com/watch?v=o9sbAZOtFAQ}.

\subsection{Task Descriptions}

In the first experiment, we implemented a four rooms task in which the ant agent needs to move from the bottom left room to the top right room.  Unlike the other two experiments, this task was relatively short horizon ($\sim 125$ primitive actions) and required minimal exploration.  In this type of task, HAC should have an advantage over HACx as HAC agents will be less likely to explore areas of the state space that do not lead to the goal.  

In the second experiment, we implement an open field reacher task, in which the ant must move from some small region in the center of the field to some distant location in one of the corners of the field.  This open field reacher environment is significantly larger than the open field reacher domain implemented in the HAC paper---the surface area of our implementation is more than 300x larger than the equivalent implementation in HAC.  (For a visual of this size difference, please see the supplementary materials, which has a figure with screenshots of the two implementations.)   Our implementation also requires significantly more exploration as the initial state space is limited to a small region in the center of the field, rather than the whole field as implemented in the HAC paper.  Our open field reacher domain requires more than 1,000 primitive actions, which is over an order of magnitude larger than the open field environment in the original HAC paper. 

In the third experiment, we implement the spiral maze task shown in Figure \ref{fig:spiral_maze}.  In this task, the agent must move from a small region in the center of the maze to a small region at the end of the maze.  This task is also significantly longer than the four rooms task, requiring over 1,000 primitive actions, or around 7.5 minutes of navigation.  Further, due to the walls as well as the small initial state and goal spaces, the task requires significant exploration.

Additionally, the task goal space in each experiment was a region of $(x,y)$ coordinates.  Given that the task goal space was the $(x,y)$ coordinate space, the RND goal space was set to the same space.  Thus, the learned and true RND functions $f$ and $\hat{f}$ map $(x,y)$ coordinates to $\mathbb{R}^d$ codes.  Also, due to the sensitivity of the ant controller, if the ant flips on its back, we automatically flip the ant back on its feet in place.
\subsection{Comparison Methods}

In these three experiments, we compared HACx to HAC, flat RND, and Curious Hierarchical Actor-Critic (CHAC) \citep{roder2020curious}.   \citet{levy2018hierarchical} showed that sample efficiency improves with additional levels so we implemented HACx, HAC, and CHAC as 4 level agents.  We implemented flat RND as a one-level HACx agent that has both a one-level goal-conditioned policy and a one-level exploration policy and uses the same RND curriculum scheme as HACx.  

CHAC is a recent effort to combine hierarchy and exploration.  CHAC attempts to help HAC agents explore by maximizing surprise.  Each level of the HAC agent learns a forward model $f_{\theta}: \mathcal{S} \times \mathcal{A} \rightarrow \mathcal{S}$ that tries to predict the next state given the prior state and action.  CHAC then uses the error or the surprise of this forward model to generative an intrinsic reward for each level similar to \citep{pathak2017curiosity}.  The intrinsic reward $R^{\text{int}}_t$ is normalized to be between $[-1,0]$ to be compatible with the extrinsic HAC reward $R^{\text{ext}}_t$.  The policy at each level attempts to maximize the sum of rewards, in which the reward at time $t$, $R_t$, is the average of $R_{\text{int}}$ and $R_{\text{ext}}$.  Given that HAC subgoals can be divided into three categories: (i) subgoals that can be achieved by the current lower level hierarchy, (ii) subgoals that cannot be achieved by the current lower policy hierarchy but could by an optimal lower level policy hierarchy, and (iii) subgoals that cannot be achieved by an optimal lower level policy hierarchy, CHAC aims for agents to output subgoals from the second category as these subgoals should lead to more surprise and they can eventually be achieved by the lower levels. 

One difficulty with CHAC is that it is not clear whether the algorithm will assign higher Q values to subgoals in the second category than subgoals in the first category that lead to frequently visited states.  HAC values subgoal actions by averaging the Bellman target Q-values (i.e., $r_t + \gamma Q(s_t,g,\pi(s_t,g))$) from the Future UMDP and Present UMDP transitions.  With CHAC, although the target Q values values from the Future UMDP transitions may be higher for subgoals in the second category than subgoals from the first category due to the inclusion of the intrinsic reward, the average target Q value that includes the subgoal testing transitions from the Present UMDP may not be.  If the subgoal that leads to a relatively new state cannot be achieved by the current lower level policy hierarchy, that subgoal may be penalized with a Present UMDP transition containing the penalty reward of $-H$.  The relatively high target Q value from the Future UMDP transition may then be outweighed by the low target Q value from the Present UMDP, resulting in a subgoal policy that prefers unsurprising subgoals from the first category.

In addition, with CHAC it is also likely that unachievable subgoals from the third category have higher Q-value estimates than subgoals from the desirable second category.  In subgoal penalty transitions, HAC uses a discount rate of 0 in order to overcome problems caused by changing next state distributions.  As a result, when CHAC's intrinsic reward is integrated into HAC, subgoals from the third category that will receive the subgoal penalty will have target Q-values of around $-H/2$, which results from averaging $R^{\text{ext}}_t=-H$ from the subgoal penalty and the intrinsic reward $R^{\text{int}}_t \approx 0$.  In long time horizon tasks, it is likely that the Q-values for second category subgoals will drop below this threshold.  Consequently, it is likely that CHAC will incentivize agents to propose unobtainable subgoals, which in turn prevents the hierarchical agent from breaking down a task into smaller subtasks.

\subsection{Results}

\begin{figure*}
    \centering
    \includegraphics[width=1\linewidth]{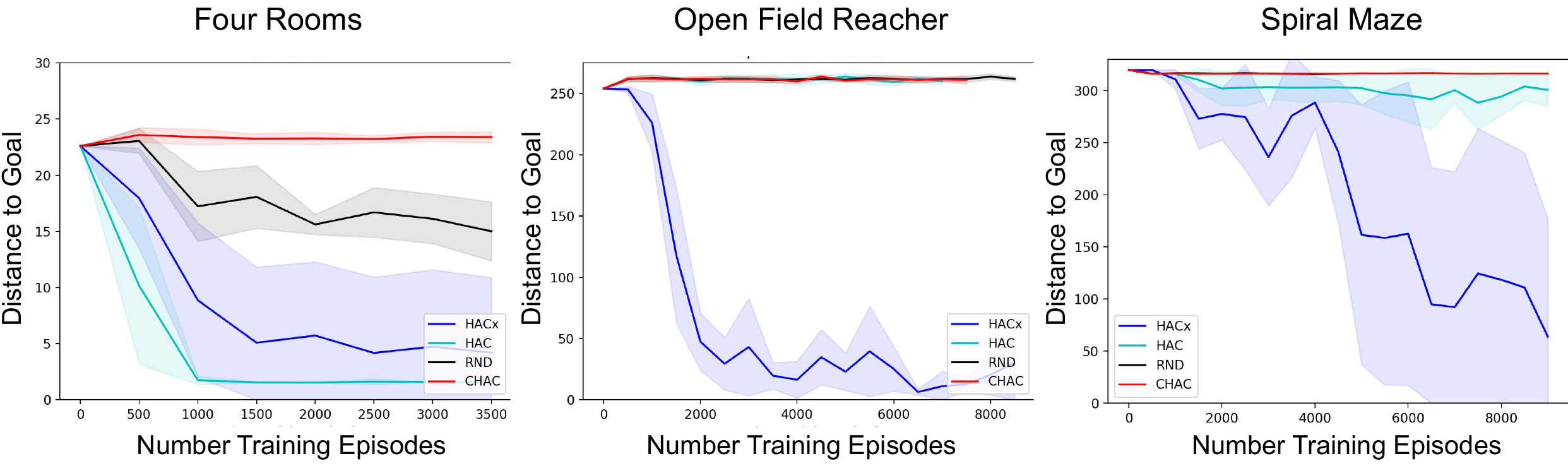}
    \caption{Performance curves for HACx and competitor methods.  In each trial in each experiment, performance was measured using the average closest distance to the goal the agent reached over 50 test episodes.  The charts show the mean and 1-standard deviation of this performance measured over 5-10 trials.  HAC slightly outperforms HACx in the easier four rooms task due to its lack of exploration, but  HACx significantly outperforms all other methods in the long time horizon domains.}
    \label{fig:results}
\end{figure*}

Figure \ref{fig:results} shows the results from the three domains.  In the ant four rooms task which required minimal exploration, HACx significantly outperformed RND and CHAC, but learned with less sample efficiency than HAC.  However, in the two more difficult tasks that required substantial long distance exploration, HACx was the only method that could consistently solve the tasks.  The other methods were only able to move a short distance away from the initial state space.  In addition, when the trained CHAC agents were visualized we found that the subgoals that were proposed were not near the reachable range.  This outcome was expected as unachievable subgoals can be encouraged by CHAC as they have a relatively high lower bound in Q-values.  

\begin{figure}
    \centering
    \includegraphics[width=1.0\linewidth]{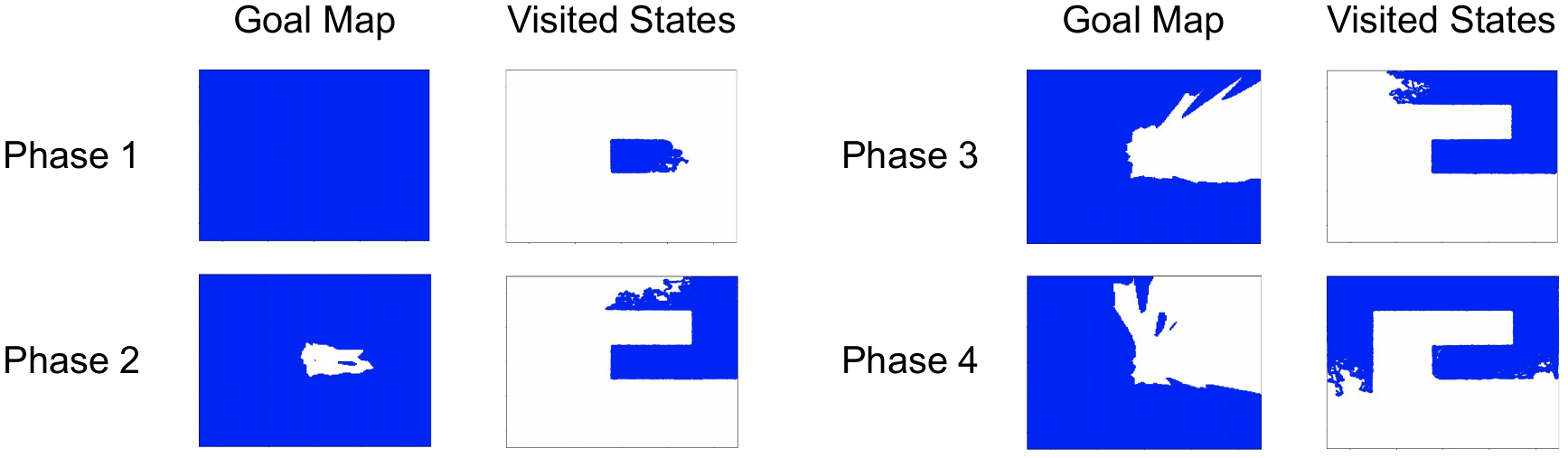}
    \caption{RND goal and state visitation maps are shown from the initial phases of the RND curriculum in the spiral maze domain.  The goal maps show in blue the states that qualify as RND goals at the beginning of a curriculum phase.  The state visitation maps show in blue the states visited in the next phase of the curriculum.}
    \label{fig:spiral_maps}
\end{figure}

Figure \ref{fig:spiral_maps} shows how the RND curriculum incentives agents to explore the environment.  The figure shows the first four phases of the RND curriculum from a trial in the spiral maze domain. The left chart in each phase row shows in blue the states that are considered as RND goal states before the phase begins.  For instance, in the top left chart, all states are blue as training has not started yet.  The right chart in each phase row shows the states that were visited in the next phase.  As the charts show, the RND goal regions encourage the agent to expand the frontier of states that have been visited, which then further shrinks the RND goal regions, motivating further exploration.  

\section{Related Work}

There is a large body of work in HRL \citep{dayan1993feudal, sutton1999between, bakker2004hierarchical, dietterich2000hierarchical,konidaris2009skill}.  Most recent methods fall into two main categories: (i) feudal-inspired \citep{dayan1993feudal} methods and (ii) skill or option-based methods \citep{sutton1999between}.

Feudal methods, like HAC, use the state space to divide up a task into a set of easier subtasks.  Compared to HACx, recent methods either (i) do not fully address the issue of nonstationary transition and rewards function or (ii) do not explicitly encourage exploration. HIRO \citep{NEURIPS2018_e6384711} does not always replace the original subgoal action with the subgoal state achieved in hindsight so HIRO may not be as effective as HAC at overcoming the nonstationary problem in HRL, which may partly explain why HAC significantly outperformed HIRO in a set of simulated robotics experiments \citep{levy2018hierarchical}.  Relay Policy Learning (RPL)  \citep{DBLP:journals/corr/abs-1910-11956} applies a similar subgoal replacement scheme as HAC, but relies on demonstrations to help agents explore.

Skill-based methods train two-level hierarchical agents.  The bottom level learns a discrete set or continuous space of skills, each of which include a policy and initiation and termination conditions for when to start and stop the skill.  The top level learns a policy over the skill space.  HACx likely has an advantage in learning lengthy policies relative to these methods because HACx can learn more than two levels of policies in parallel.  Deep Skill Chaining (DSC) \citep{Bagaria2020Option} concatenates skills backwards one at a time from a goal state to an initial state, but is not able to learn the skills and the policy over the skills simultaneously. Diversity Is All You Need (DIAYN) \citep{eysenbach2018diversity} uses a two stage process to train its agents.  The first stage learns a discrete set of diverse options and then in the second stage learns a policy over these options to maximize reward.  HIDIO \citep{zhang2021hierarchical} does learn a continuous space of diverse options and a policy over the options at the same time.  But the policy over options likely needs to wait for the diverse skills to stabilize before the policy over options can be trained effectively.  Both DIAYN and HIDIO do explicitly encourage exploration by including a term in the objective function that encourages the skill policies to maximize entropy.  We believe HACx is likely more effective at exploring because (i) HACx directly incentivizes agent to reach new states and (ii) its behavioral subgoal policies also act with relativley high entropy but they act at larger time scales so the range of states that can be realistically achieved is larger.

\section{Conclusion}

We propose a new algorithm, HAC Explore, that can help agents explore over long distances.  The method combines the RND exploration bonus technique and the HAC hierarchical approach.  RND actively encourages the agent to explore new states, while the HAC component accelerates the exploration process by dividing up the task of reaching distant new states.  The combined approach significantly outperformed HAC, RND, and another integrated hierarchical and exploration method in a pair of challenging domains that required over 1,000 primitive actions to solve.

\bibliography{refs}
\bibliographystyle{unsrtnat}



\end{document}